\documentclass[letterpaper]{article} % DO NOT CHANGE THIS
\usepackage{aaai25}  % DO NOT CHANGE THIS
\usepackage{times}  % DO NOT CHANGE THIS
\usepackage{helvet}  % DO NOT CHANGE THIS
\usepackage{courier}  % DO NOT CHANGE THIS
\usepackage[hyphens]{url}  % DO NOT CHANGE THIS
\usepackage{graphicx} % DO NOT CHANGE THIS
\usepackage{natbib}  % DO NOT CHANGE THIS AND DO NOT ADD ANY OPTIONS TO IT
\usepackage{caption} % DO NOT CHANGE THIS AND DO NOT ADD ANY OPTIONS TO IT
\usepackage{graphicx}
\usepackage{amsmath,amssymb}
\usepackage{newtxmath}
\usepackage{color}
\usepackage[noend]{algpseudocode}
\usepackage{algorithmicx,algorithm}
\usepackage{url}
\usepackage{bigstrut,multirow,rotating}
\usepackage{threeparttable}
\usepackage{booktabs}
\usepackage{array}
\usepackage{pifont}
\usepackage{courier}
\newcommand{\tabincell}[2]{\begin{tabular}{@{}#1@{}}#2\end{tabular}}

\usepackage[table,xcdraw]{xcolor}

\definecolor{lightblue}{rgb}{0.761,0.894,0.937} % RGB(194, 228, 239)
\definecolor{white}{rgb}{0.965,0.973,0.988} % White
\definecolor{lightpurple}{rgb}{0.914,0.870,0.933} % RGB(233, 222, 299)

\usepackage{newfloat}
\usepackage{listings}
\DeclareCaptionStyle{ruled}{labelfont=normalfont,labelsep=colon,strut=off} % DO NOT CHANGE THIS
\floatstyle{ruled}
\newfloat{listing}{tb}{lst}{}
\floatname{listing}{Listing}
\title{ChangeDiff: A Multi-Temporal Change Detection Data Generator with Flexible Text Prompts via Diffusion Model}
\author{
    Qi Zang\textsuperscript{\rm 1},
    Jiayi Yang\textsuperscript{\rm 1},
    Shuang Wang\textsuperscript{\rm 1}\thanks{Corresponding author},
    Dong Zhao\textsuperscript{\rm 1},
    Wenjun Yi\textsuperscript{\rm 1},
    Zhun Zhong\textsuperscript{\rm 2}
}
\affiliations{
    \textsuperscript{\rm 1}School of Artificial Intelligence, Xidian University, China\\
    \textsuperscript{\rm 2}School of Computer Science and Information Engineering, Hefei University of Technology, China
}

\usepackage{bibentry}
\begin{document}

\maketitle

\begin{abstract}
Data-driven deep learning models have enabled tremendous progress in change detection (CD) with the support of pixel-level annotations. However, collecting diverse data and manually annotating them is costly, laborious, and knowledge-intensive. Existing generative methods for CD data synthesis show competitive potential in addressing this issue but still face the following limitations: 1) difficulty in flexibly controlling change events, 2) dependence on additional data to train the data generators, 3) focus on specific change detection tasks. 
To this end, this paper focuses on the semantic CD (SCD) task and develops a multi-temporal SCD data generator ChangeDiff by exploring powerful diffusion models.
ChangeDiff innovatively generates change data in two steps: first, it uses text prompts and a text-to-layout (T2L) model to create continuous layouts, and then it employs layout-to-image (L2I) to convert these layouts into images. 
Specifically, we propose multi-class distribution-guided text prompts (MCDG-TP), allowing for layouts to be generated flexibly through controllable classes and their corresponding ratios. Subsequently, to generalize the T2L model to the proposed MCDG-TP, a class distribution refinement loss is further designed as training supervision.
%For the former, a multi-classdistribution-guided text prompt (MCDG-TP) is proposed to complement via controllable classes and ratios. To generalize the text-to-image diffusion model to the proposed MCDG-TP, a class distribution refinement loss is designed as training supervision. For the latter, MCDG-TP in three modes is proposed to synthesize new layout masks from various texts. 
Our generated data shows significant progress in temporal continuity, spatial diversity, and quality realism, empowering change detectors with accuracy and transferability.
The code is available at \textcolor{red!80!black}{https://github.com/DZhaoXd/ChangeDiff}.
\end{abstract}


\section{Introduction}
Change detection (CD), a key Earth observation task, employs bitemporal remote sensing data to gain a dynamic understanding of the Earth's surface, producing pixel-wise change maps for ground objects \cite{feranec2007corine,chen2013multi,kadhim2016advances}.
In recent years, data-driven deep learning models have provided promising tools for CD and achieved remarkable progress \cite{lei2019multiscale,arabi2018optical,dong2018local}. These advancements rely on large-scale, high-quality pixel-level annotations. However, building such a dataset poses a significant challenge because collecting diverse data and manually annotating them is costly, labor-intensive, and requires expert intervention. As a result, these challenges unsurprisingly restrict the size of existing public CD datasets, compared to general-purpose vision datasets such as ImageNet \cite{deng2009imagenet}.

\begin{table}[tbp]
  \centering
   \renewcommand{\arraystretch}{1.05}
  \resizebox{\linewidth}{!}{
\begin{tabular}{c|cccc}
\toprule
     Methods & Text Control & Layout Diversity & w/o Extra Seg. Data  & SCD Task \\
\hline
IAug \cite{chen2021adversarial} & \ding{55}    & \ding{55}    & \ding{51}   & \ding{55} \\
ChangeStar \cite{zheng2021change} & \ding{55}    & \ding{55}    & \ding{55}    & \ding{55} \\
Self-Pair \cite{seo2023self} & \ding{55}    & \ding{55}    & \ding{55}    & \ding{55} \\ 
Changen  \cite{zheng2023scalable} & \ding{55}    & \ding{55}    & \ding{55}    & \ding{51} \\ 
\hline
\cellcolor{lightblue} ChangeDiff (Ours) &\cellcolor{lightblue!50!white} \ding{51} &\cellcolor{white} \ding{51} &\cellcolor{lightpurple!50!white} \ding{51} &\cellcolor{lightpurple} \ding{51} \\
\bottomrule
\end{tabular}}
 \setlength{\abovecaptionskip}{0.05 cm}
  \caption{Comparison of data synthesis method in change detection regarding functionality, data support, and application tasks. Our ChangeDiff shows more practical and strong functions and application scenarios.}
  \label{comp_gen}
  \vspace{-0.5cm}
\end{table}

\begin{figure*}[tbp]
    \begin{center}
    \centering 
\includegraphics[width=1\textwidth]{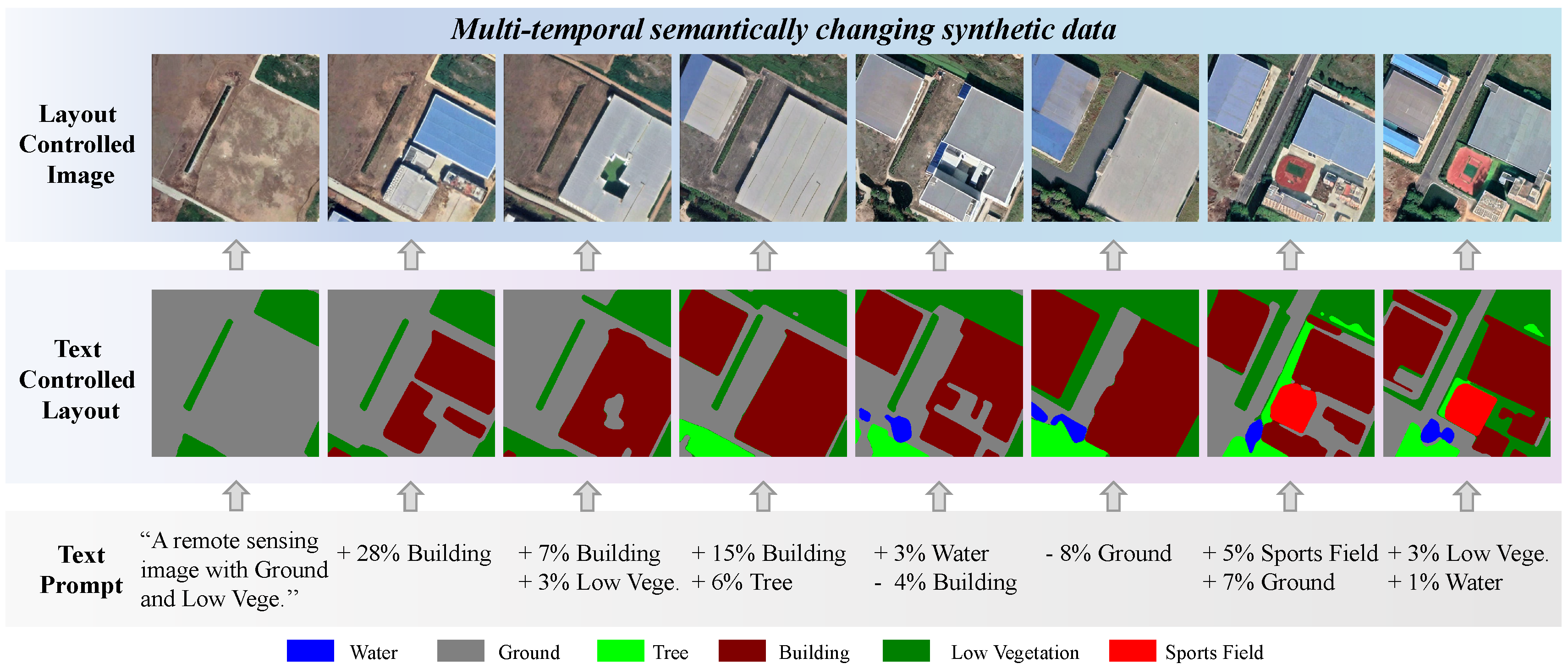}
    \end{center}
\setlength{\abovecaptionskip}{-0.15 cm}
%    \captionsetup{font={small}}
    \caption{Multi-temporal semantic change synthetic data synthesized by our ChangeDiff, which is trained on the sparsely labeled semantic change detection \textit{SECOND} \cite{yang2021asymmetric} dataset. 
    It describes an \textit{area under construction, where man-made facilities are gradually being completed.}
    ChangeDiff takes text prompts as input, generates semantic events, and specifies changes in a controllable manner by modifying the text prompts.
}
    \label{overview}
%\vspace{-0.2cm}
\end{figure*}

To alleviate high demand for data annotation, data synthesis has emerged as an alternative solution with promising application potential. Currently, a few synthesis techniques for binary CD (\emph{e.g., building variations}) have been studied, categorized into two mainstreams: data augmentation-based and data generation-based methods. 
In the former, IAug \cite{zheng2021change} and Self-Pair \cite{seo2023self} use copy-paste and image inpainting techniques, pasting instances or patches from other regions onto target images to simulate building changes. However, the inconsistency between pasted areas and backgrounds makes it challenging to create realistic scene changes. 
In the latter, Changen \cite{zheng2023scalable} introduces a generic probabilistic graphical model to generate continuous change pairs, improving the realism of synthetic images. 
However, Changen still relies on copy-paste operation of the image mask (semantic layout) to create changes, making it difficult to flexibly control change events. Additionally, the mask-based copy-paste is not easily applicable to semantic CD (SCD) task due to the lack of complete masks. Moreover, it requires additional segmentation data to train the probabilistic model, limiting transferability to specific target data. A detailed comparison of these methods is provided in Table 1.

Recently, driven by latent diffusion models \cite{rombach2022high}, generative models have reached a new milestone\cite{khanna2023diffusionsat}.  Stable Diffusion \cite{podell2023sdxl} and DALL-E2 \cite{ramesh2022hierarchical} introduce large-scale pretrained text-to-image (T2I) diffusion models that can generate high-quality images matching textual descriptions. 
Furthermore, advanced work, \emph{e.g.}, ControlNet \cite{zhang2023adding},  has shown that by incorporating fine-grained controls, such as semantic layouts, textures, or depth, T2I models can be adapted into layout-to-image (L2I) models, allowing for more flexible generation of images matching input layouts. 
\textit{This inspires us to question whether advanced T2I and L2I models can be applied to CD data synthesis to enhance CD tasks.}

Through our analysis, we identify the core challenge as: \textit{how to construct continuous change events using input text}, which arises from the following aspects:
$\textcircled{1}$ In CD tasks, especially SCD, semantic annotations are incomplete (sparse), with only the semantic classes of changed areas being labeled. This makes it difficult to create precise text-to-image mappings and train T2I or L2I models.
$\textcircled{2}$ Common used text prompts, such as ``{\fontfamily{pcr}\selectfont A remote sensing image of \{classname\}}", are inadequate for present continuous change events. We need to develop suitable text prompts tailored for CD tasks.
$\textcircled{3}$ Text-to-image mapping does not provide spatial semantics of the generated images, rendering the synthesized image pairs for downstream supervised training.

In this paper, we explore the potential of T2I and L2I models for SCD tasks and develop a novel multi-temporal SCD data generator without requiring paired images or external datasets, coined as \textbf{ChangeDiff}. 
To address the core challenges, ChangeDiff innovatively divides the change data generation into two steps: 1) it uses carefully designed text prompts and text-to-layout (T2L) diffusion models to generate continuous layouts; 2) it employs layout-to-image (L2I) diffusion model to transform these layouts into continuous time-varying images, as illustrated in Fig.~\ref{overview}.
Specifically, we innovatively develop a text-to-layout (T2L) generation model using multi-class distribution-guided text prompt (MCDG-TP) as input to generate layout flexibly. Our MCDG-TP translates the layout into semantic class distributions via text, \emph{i.e.}, each class with a class ratio, which offers a simple yet powerful control over the scene composition. Meanwhile, the ingredients of MCDG-TP differ from the text prompts used for pre-training the T2I model. This difference prevents the T2I model from generalizing to texts with arbitrary compositions, as text is the sole driver of optimization during noise prediction. To address this, we design a class distribution refinement loss to train our T2L model. With the trained T2L model, sparse layouts enable completion by inputting text with amplified class ratios; then, taking the completed layout as a reference, time-varying events can be simulated via MCDG-TP in three modes: ratio reshaper, class expander, and class reducer. Subsequently, the fine-tuned L2I model synthesizes new images aligned with the simulated layout masks. The data generated by our ChangeDiff shows significant progress in temporal continuity, spatial diversity, and quality realism. The main contributions of this paper are five-fold:

\begin{itemize}
\item To the best of our knowledge, we are the first to explore the potential of diffusion models for the SCD task and develope the first SCD data generator ChangeDiff.
\item We propose a multi-class distribution-guided text prompt (MCDG-TP), using controllable classes and ratios, to complement sparse layout masks.
\item We propose a class distribution refinement loss as training supervision to generalize the text-to-image diffusion model to the proposed MCDG-TP.
\item We propose MCDG-TP in three modes to simulate time-varying events for synthesizing new layout masks and corresponding images.
\item Extensive experiments demonstrate that our high-quality generated data is beneficial to boost the performance of existing change detectors and has better transferability.
\end{itemize}

\section{Related Work}

\noindent \textbf{Binary \& Semantic Change Detection \& Text-guided Diffusion Model.} Please refer to the \textcolor{red!80!black}{Appendix A}.

\noindent \textbf{Data Synthesis in Change Detection.} Currently, there are several advanced data synthesis methods for the change detection task.
ChangeStar \cite{zheng2021change} and Self-Pair \cite{seo2023self} employ simple copy-paste operations, pasting patches from other regions onto the target image to simulate changes. However, artifacts introduced by the paste operation and the inconsistency between foreground and background make it challenging to create realistic scene changes. 
IAug \cite{chen2021adversarial} uses a generative model (GAN) to synthesize changed objects, but its building-specific modeling approach limits its generalization to diverse scenes. 
Changen \cite{zheng2023scalable} proposes a generic probabilistic graphical model to represent continuous change events, enhancing the realism of synthetic images. 
Its latest version, Changen2 \cite{zheng2024changen2}, introduces a diffusion transformer model, further improving generation quality. 
However, it relies on additional segmentation data to train the probabilistic model, making it unsuitable for direct data augmentation in change detection tasks.

Our ChangeDiff does not rely on additional segmentation data, simplifying its integration into existing workflows. Besides, ChangeDiff supports text control, enabling users to specify the generated changes precisely. Furthermore, ChangeDiff can synthesize diverse and continuous layouts, which is crucial for improving the transferability of synthetic data.

\section{Method}
\begin{figure*}[t]
    \setlength{\abovecaptionskip}{-0.15cm}
    \begin{center}
    \centering 
    \includegraphics[width=0.91\textwidth, height=0.37\textwidth]{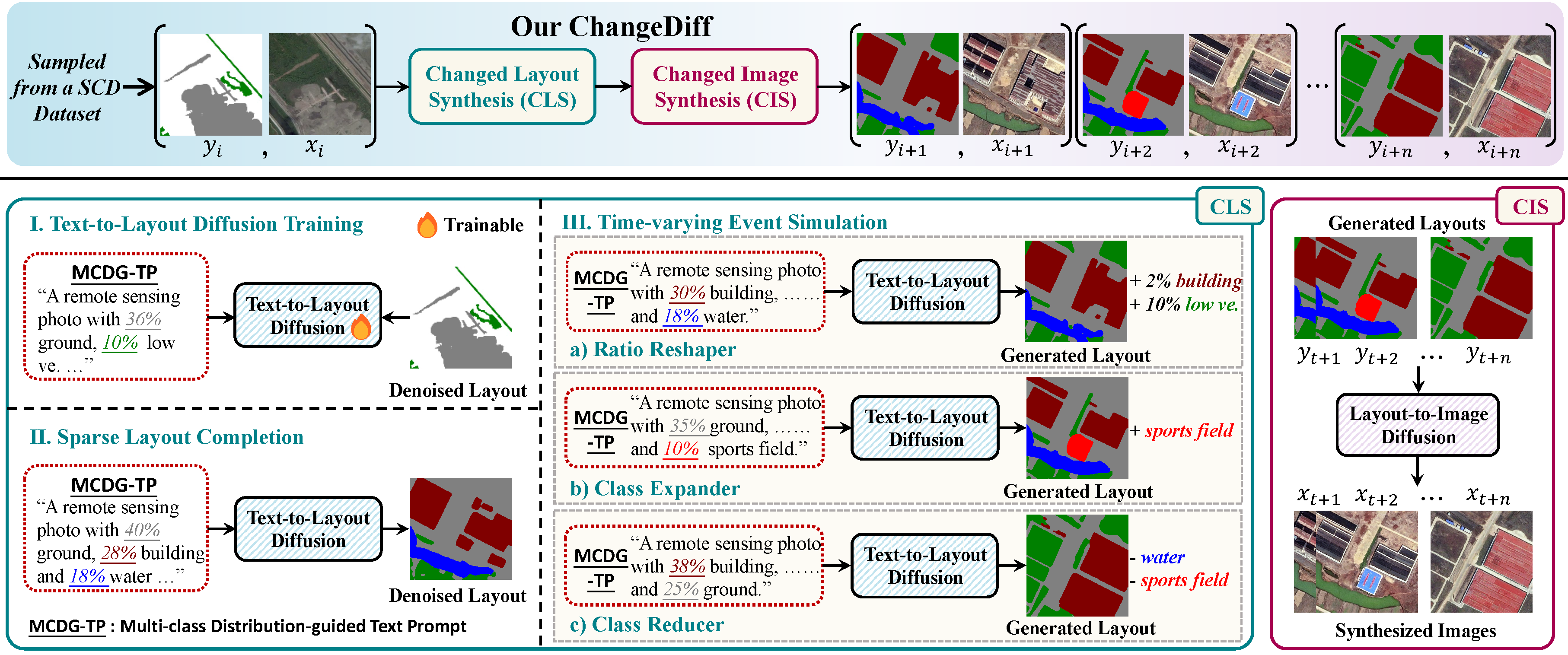} 
    \end{center}
    \caption{An overview of the proposed ChangeDiff, which consists of two components, \emph{i.e.}, changing layout synthesis (CLS) and changing image synthesis (CIS). In the CLS: I. The text-to-layout (T2L) model is fine-tuned on the target data via MCDG-TP; II. Text with amplified class ratios is fed into the fine-tuned T2L model to generate the completed layouts; III. Taking the completed layout as a reference, texts with different class compositions are fed into the fine-tuned T2L model to synthesize temporally changed layouts. In the CIS: New images aligned with the changed layouts are synthesized via a layout-to-image model fine-tuned to the target data.} 
    \label{pipeline}
    \vspace{-0.2cm}
\end{figure*}

Given a single-temporal image $x\in \mathbb{R}^{H \times W \times 3}$ and its sparsely-labeled (only the change area) semantic layout $y\in \mathbb{R}^{H \times W}$ in the semantic change detection (SCD) dataset, our SCD data generator ChangeDiff aims to simulate temporal changes via diffusion models conditioned on diverse completed semantic layouts. The pipeline of ChangeDiff is shown in Fig. \ref{pipeline}. 
Overall,  ChangeDiff consists of the text-to-layout (T2L) diffusion model for changing layout synthesis and the layout-to-image (L2I) diffusion model for changing image synthesis. 
In the next part, we first introduce the detailed design of T2L and L2I diffusion models.
Then, we discuss how to flexibly encode semantic layouts into the text prompt.  
Lastly, we introduce how to synthesize complete and diverse layouts via text prompts.

\subsection{Preliminary}
Both T2L and L2I models are built on the latent diffusion model (LDM) \cite{rombach2022high}, widely used in conditional generation tasks for its powerful capabilities. LDM learns the data distribution by constructing a $T$-step denoising process in the latent space for the normally distributed variables added to the image $x\in \mathbb{R}^{H \times W \times 3}$. Given a noisy image $x_t$ at time step $t\in \{1,\dots,T\}$, the denoising function $\epsilon_\theta$ parameterized by a U-Net \cite{ronneberger2015u} learns to predict the noise $\epsilon$ to recover $x$.
\begin{equation}
\mathcal{L}_{LDM} = \mathbb{E}_{z_0,t,\epsilon \sim \mathcal{N}(0,1)} \| \epsilon - \epsilon_{\theta}(z_t, t, \tau_\theta(\mathbf{T})) \|^{2},  \label{ldm}
\end{equation}
where $z=\varepsilon_\theta(x)$ is the encoding of the image in latent space. $\tau_\theta$ is a pre-trained text encoder \cite{radford2021learning} that enables cross-attention between the text $\mathbf{T}$ and $z_t$ to steer the diffusion process.

\subsubsection{T2L Model.} 
Given a semantic layout $y$ where pixel values are category IDs, we encode $y$ as a three-channel color map $\mathbf{C}_{y}$ in RGB style, and each color indicates a category. This enables leveraging the pre-trained LDM model to generate semantic layouts from input text prompts $\mathbf{T}$.

\subsubsection{L2I Model.} Recent works like ControlNet \cite{zhang2023adding} propose conditioning on both semantic layout $y$ and text prompt $\mathbf{T}$ to synthesize images aligned well with specific semantic layouts. Following this, we adopt the ControlNet structure, which adds a trainable side network to the LDM model for encoding semantic layout $y$.

\subsection{Flexible Text Prompt as Condition}
We explore encoding semantic layouts via text prompts to utilize the T2L model for completing sparse layouts and generating diverse layouts.

\subsubsection{Semantic Layout Translation.} 
A semantic layout $y$ can be decomposed into two components: category names $\{n_j\}$ and their corresponding connected areas $\{a_j\}$. The names can be naturally encoded by filling in the corresponding textual interpretation into $\mathbf{T}$, but connected areas can not since the pixels within them are arranged consecutively. To discretize, we partition each pixel into cells at distinct coordinate points. Cells with the same category ID can be count-aggregated into a unique class ratio, quantitatively characterizing the corresponding $\{a_j\}$ for insertion into the encoding vocabulary of $\mathbf{T}$. Formally, given a semantic layout $y$, the connected areas $a_j$ of the $j$-th class are represented by the class ratio $R_j$ as,
\begin{equation}
R_j = \sum_{hw}^{HW}[\vmathbb{1}(y_{hw}^{}=j)]/HW.
\end{equation}
$\vmathbb{1}(\cdot)$ is an indicator function, which is $1$ if the category ID is $j$, otherwise it is $0$. Then, a certain category $j$ in the semantic layout $y$ is encoded as a phrase $(n_j,R_j)$ with two tokens.

\subsubsection{Text Prompt Construction.} 
With the phrases encoding multiple categories, we serialize them into a single sequence to generate text prompts. All phrases will be sorted randomly. Specifically, we adopt a template to construct the text prompt, ``{\fontfamily{pcr}\selectfont A remote sensing photo with \{class distributions\}}", where class distributions = ``$\dots(n_{j-1},R_{j-1})(n_j,R_j)(n_{j+1},R_{j+1})\dots$". We term this as multi-class distribution-guided text prompt (MCDG-TP).

\subsection{Class Distribution Refinement Loss}
The loss function $\mathcal{L}_{LDM}$ used to pre-train the LDM model only optimizes the denoising function $\epsilon_\theta$ (See Eq. (\ref{ldm})), enabling it to predict noise and thus recover the clean image. There is no explicit constraint between the embeddings $\tau_\theta(\mathbf{T})$ of text prompts and noisy images $z_t$ in the latent space. As a result, when provided with refined text prompts during inference, the learned denoising function $\epsilon_\theta$ may fail to generate the corresponding object composition. To better understand the text prompt $\mathbf{T}$, we design a class distribution refinement loss $\mathcal{L}_{CDR}$ that explicitly supervises the cross-attention map between $\tau_\theta(\mathbf{T})$ and $z_t$ during training. Our $\mathcal{L}_{CDR}$ assists the model in capturing information from the budget ratio and spatial location.
For a cross-attention map at any layer $m\in M$, the $\mathcal{L}_{CDR}$ is defined as follows,
\begin{equation}
\mathcal{L}_{CDR} = \underbrace{\mathcal{L}_{RAT}}_{\text{ratio}} + \underbrace{\mathcal{L}_{SPA}}_{\text{spatial location}}.
\end{equation}

For the ratio, the cross-attention maps $\mathcal{A}_{class}$ of objects are weighted by that of their corresponding ratios $\mathcal{A}_{ratio}$ to define a combined map $\mathcal{A}_{com}=\mathcal{A}_{class}\cdot\mathcal{A}_{ratio}$. $\mathcal{A}_{com}$ is taken as the constraint target in $\mathcal{L}_{RAT}$,
\begin{equation}
\mathcal{L}_{RAT} = \frac{1}{J}\sum_{j}^{J}\left\lvert\frac{\sum_{hw}^{H_mW_m}[\vmathbb{1}(\mathcal{A}_{(com,j)}^{hw}\textgreater 0)]}{H_mW_m} - R_{j}^{}\right\rvert.
\end{equation}
$\mathcal{A}_{(com,j)}^{hw}$ is the intersection of class activations in the generated features and the GT. $H_m\times W_m$ is the map size of the $m$-th layer. 
This formula enforces that the cross-attention activations from the diffusion model align with the true class ratio \( R_{j} \) for any class \( j \). 

For the spatial location, we acquire a binary segmentation map $\mathcal{M}_j$ for each object in $\mathbf{T}$ from its respected semantic layout $y$, providing the ground truth distribution. $\mathcal{L}_{SPA}$ aggregates pixel-level activation values of objects and encourages their even distribution,
\begin{equation}
\mathcal{L}_{SPA} = \frac{1}{JH_mW_m}\sum_{j}^{J}\sum_{hw}^{H_mW_m}(\mathcal{A}_{(class,j)}^{hw} - \mathcal{M}_{j}^{m})_{}^{2}.
\end{equation}
$\mathcal{A}_{(class,j)}^{hw}$ is the cross-attention map from the diffusion model, which means activations of category words on generated image features. $\mathcal{M}_{j}^{m}$ is formed through bilinear interpolation followed by binarization to match the resolution of the map of the $m$ -th layer.
This formula enforces spatial alignment of activations with the true response $\mathcal{M}_{j}^{m}$.

\subsection{Changing Layout Synthesis}
To synthesize the changed images, reasonable and diverse layout synthesis in the temporal dimension is required as a semantic guide.

\subsubsection{T2L Model Training.}
Given the target SCD dataset, we use the [text prompt $\mathbf{T}_y$, color map $\mathbf{C}_y$] training pairs to fine-tune the T2L model, where $\mathbf{T}_y$ is the corresponding text generated via our MCDG-TP for the color map. With the proposed loss $\mathcal{L}_{CDR}$ and the original loss $\mathcal{L}_{LDM}$, the T2L model is supervised during fine-tuning as follows,
\begin{equation}
\mathcal{L}_{\mathrm{ChangeDiff}} = \mathcal{L}_{LDM} + \sum_{m}^{M}\mathcal{L}_{CDR}^{m}.
\end{equation}

\subsubsection{Sparse Layout Completion.}
Since generating a changed image requires a complete layout, it is necessary to complete the sparse layout to serve as a reference for synthesizing the changed layout. To obtain the completed layout, we input text prompt $\mathbf{T}_c$ with amplified class ratios and random noise $z_c$ sampled from $z\sim \mathcal{N}(0,1)$ into the fine-tuned T2L model. By varying $\mathbf{T}_c$ and $z_c$, various reference color maps $\mathbf{C}_i$ with different object compositions can be obtained.

\subsubsection{Time-varying Event Simulation.}
With an arbitrary reference layout and its corresponding [$\mathbf{T}_c$, $z_c$], we input noise sampled following $z_c$ and varied text into the fine-tuned T2L model to simulate real-world time-varying events. For the varied text, we construct MCDG-TP in three modes, a) ratio reshaper $\mathbf{T}_s$: randomly change the ratio of each class in $\mathbf{T}_c$; b) class expander $\mathbf{T}_e$: randomly create new classes into $\mathbf{T}_c$; c) class reducer $\mathbf{T}_d$: randomly remove certain classes from $\mathbf{T}_c$. Diverse changed color maps $\{\mathbf{C}_{i+1},\dots,\mathbf{C}_{i+n}\}$ in spatiality can be obtained, and then we can get the changed layout masks $\{y_{i+1},\dots,y_{i+n}\}$ via a learning-free projection function $f_{color\rightarrow mask}: \mathbb{R}^{H \times W \times 3}\rightarrow \mathbb{R}^{H \times W \times J}$, \textit{i.e.}, maps RGB values to mask space by simple color matching, \textit{e.g.}, mapping red pixel ([255, 0, 0]) to class id 1.

\subsection{Changing Image Synthesis}
Conditioned on the synthesized color maps $\{\mathbf{C}_{i},\dots,\mathbf{C}_{i+n}\}$ and texts $\{\mathbf{T}_{i},\dots,\mathbf{T}_{i+n}\}$, we use the fine-tuned L2I model to synthesize images $\{x_{i},\dots,x_{i+n}\}$ aligned with the given layouts. During synthesis, we randomly sample a noise $z_i$ from $z\sim \mathcal{N}(0,1)$ to obtain a reference image $x_i$. Starting from $x_i$, the semantic content of images over time should remain within a certain similarity range. To this end, the input noises $\{z_{i+1},\dots,z_{i+n}\}$ for the changed images is sampled via a stitching mechanism, which is formulated by,
\begin{equation}
z_{i+1} = \alpha z_i + (1-\alpha) z_{r(r\neq i)}.
\end{equation}
$z_{r(r\neq i)}$ is arbitrary noise sampled from $z\sim \mathcal{N}(0,1)$. Proportional injection of $z_{r(r\neq i)}$ ensures the semantic content of the synthesized image conforms to realistic and reasonable changes, \emph{i.e.}, be temporally continuous. At this point, the synthesized images can be paired with layout masks to form new large-scale training samples with dense annotations, $\{(x_{i},y_{i}),\dots,(x_{i+n},y_{i+n})\}$.

\begin{table}[tbp]
  \centering
   \renewcommand{\arraystretch}{1.05}
  \resizebox{\linewidth}{!}{
\begin{tabular}{cccccccccccc}
\toprule
\multicolumn{12}{c}{Train on 5\% SECOND Train Set} \\
\midrule
Methods & Params. (M) & \textcolor[rgb]{ .749,  .561,  0}{OA} & \textcolor[rgb]{ .749,  .561,  0}{Score} & \textcolor[rgb]{ .749,  .561,  0}{mIoU} & \textcolor[rgb]{ .749,  .561,  0}{Sek} & \textcolor[rgb]{ .749,  .561,  0}{$\mathrm{F_{scd}}$} & \textcolor[rgb]{ .749,  .561,  0}{Kappa} & \textcolor[rgb]{ .184,  .459,  .71}{IoU} & \textcolor[rgb]{ .184,  .459,  .71}{F1} & \textcolor[rgb]{ .184,  .459,  .71}{Rec.} & \textcolor[rgb]{ .184,  .459,  .71}{Pre.} \\
\midrule
SSCDL \cite{ding2022bi} & 13.22 & \textbf{83.9}  & 26.9  & 65.4  & 10.5  & 50.8  & 54.9  & 46.4  & 63.4  & 61.8  & \textbf{65.0} \\
\rowcolor[rgb]{ .851,  .851,  .851} SSCDL + Ours &       & 83.5 & \textbf{29.0} & \textbf{66.3} & \textbf{13.0} & \textbf{53.0} & \textbf{56.8} & \textbf{48.7} & \textbf{65.5} & \textbf{69.1} & 62.3  \\
\midrule
BiSRNet \cite{ding2022bi} & 13.31 & 83.6  & 26.1  & 64.8  & 9.6   & 49.5  & 53.9  & 45.4  & 62.5  & 60.7  & \textbf{64.3} \\
\rowcolor[rgb]{ .851,  .851,  .851} BiSRNet + Ours &       & \textbf{83.8} & \textbf{28.5} & \textbf{66.3} & \textbf{12.3} & \textbf{52.4} & \textbf{56.7} & \textbf{48.3} & \textbf{65.2} & \textbf{66.5} & 63.9  \\
\midrule
TED \cite{ding2024joint} & 14.1  & 83.9  & \textbf{27.1}  & 65.7  & 10.5  & 50.3  & 55.4  & 46.8  & 63.8  & 62.1  & 65.5  \\
\rowcolor[rgb]{ .851,  .851,  .851} TED + Ours &       & \textbf{84.7} & 26.7 & \textbf{66.6} & \textbf{12.4} & \textbf{53.0} & \textbf{56.8} & \textbf{48.0} & \textbf{64.9} & \textbf{62.9} & \textbf{66.9} \\
\midrule
A2Net \cite{li2023lightweight} & 3.93  & \textbf{83.8}  & 27.1  & 66.1  & 10.4  & 49.7  & 56.1  & 47.5  & 64.4  & 63.3 & \textbf{65.5}  \\
\rowcolor[rgb]{ .851,  .851,  .851} A2Net + Ours &       & 83.7 & \textbf{28.0} & \textbf{66.2} & \textbf{11.6} & \textbf{51.4} & \textbf{56.5} & \textbf{48.1} & \textbf{65.0} & \textbf{66.0}  & 63.9 \\
\midrule
SCanNet \cite{ding2024joint} & 17.81 & 82.8  & 27.4  & 65.1  & 11.3  & 51.1  & 54.8  & 47.0  & 64.0  & \textbf{67.6}  & 60.7 \\
\rowcolor[rgb]{ .851,  .851,  .851} SCanNet + Ours &       & \textbf{85.4} & \textbf{29.6} & \textbf{67.1} & \textbf{13.6} & \textbf{54.2} & \textbf{57.6} & \textbf{48.5} & \textbf{65.3} & 61.9 & \textbf{69.1}  \\
\midrule
\multicolumn{12}{c}{Train on 20\% SECOND Train Set} \\
\midrule
Methods & Params. (M)  & \textcolor[rgb]{ .749,  .561,  0}{OA} & \textcolor[rgb]{ .749,  .561,  0}{Score} & \textcolor[rgb]{ .749,  .561,  0}{mIoU} & \textcolor[rgb]{ .749,  .561,  0}{Sek} & \textcolor[rgb]{ .749,  .561,  0}{$\mathrm{F_{scd}}$} & \textcolor[rgb]{ .749,  .561,  0}{Kappa} & \textcolor[rgb]{ .184,  .459,  .71}{IoU} & \textcolor[rgb]{ .184,  .459,  .71}{F1} & \textcolor[rgb]{ .184,  .459,  .71}{Rec.} & \textcolor[rgb]{ .184,  .459,  .71}{Pre.} \\
\midrule
A2Net \cite{li2023lightweight} & 3.93  & 84.9  & 31.7  & 68.9  & 15.8  & 55.9  & 60.9  & 52.3  & 68.7  & \textbf{71.6}  & 66.0  \\
\rowcolor[rgb]{ .851,  .851,  .851} A2Net + Ours &       & \textbf{86.1} & \textbf{32.5} & \textbf{69.6} & \textbf{16.6} & \textbf{57.3} & \textbf{61.9} & \textbf{52.8} & \textbf{69.1} & 68.1 & \textbf{70.1} \\
\midrule
SCanNet \cite{ding2024joint} & 17.81 & 84.8  & 31.9  & 68.6  & 16.2  & 56.5  & 60.4  & 51.9  & 68.4  & \textbf{72.0} & 65.1  \\
\rowcolor[rgb]{ .851,  .851,  .851} SCanNet + Ours &       & \textbf{86.9} & \textbf{33.8} & \textbf{70.1} & \textbf{18.2} & \textbf{59.1} & \textbf{62.6} & \textbf{53.2} & \textbf{69.4} & 66.5  & \textbf{72.6} \\
\midrule
\multicolumn{12}{c}{ Train on 100\% SECOND Train Set} \\
\midrule
Methods & Params. (M)  & \textcolor[rgb]{ .749,  .561,  0}{OA} & \textcolor[rgb]{ .749,  .561,  0}{Score} & \textcolor[rgb]{ .749,  .561,  0}{mIoU} & \textcolor[rgb]{ .749,  .561,  0}{Sek} & \textcolor[rgb]{ .749,  .561,  0}{$\mathrm{F_{scd}}$} & \textcolor[rgb]{ .749,  .561,  0}{Kappa} & \textcolor[rgb]{ .184,  .459,  .71}{IoU} & \textcolor[rgb]{ .184,  .459,  .71}{F1} & \textcolor[rgb]{ .184,  .459,  .71}{Rec.} & \textcolor[rgb]{ .184,  .459,  .71}{Pre.} \\
\midrule
SSCDL \cite{ding2022bi} & 13.22 & 87.3  & 36.0  & 71.9  & 20.6  & 60.2  & 65.4  & 56.1  & 71.9  & 68.3  & \textbf{75.8}  \\
\rowcolor[rgb]{ .851,  .851,  .851} SSCDL + Ours &       & \textbf{88.2} & \textbf{38.1} & \textbf{73.3} & \textbf{23.0} & \textbf{63.3} & \textbf{67.4} & \textbf{58.1} & \textbf{73.5} & \textbf{71.6} & 75.5 \\
\midrule
BiSRNet \cite{ding2022bi} & 13.31 & 87.4  & 36.2  & 71.9  & 20.9  & 60.8  & 65.3  & 56.0  & 71.8  & 68.2  & 75.8  \\
\rowcolor[rgb]{ .851,  .851,  .851} BiSRNet + Ours &       & \textbf{88.4} & \textbf{38.4} & \textbf{73.4} & \textbf{23.4} & \textbf{63.7} & \textbf{67.6} & \textbf{58.2} & \textbf{73.5} & \textbf{70.6} & \textbf{76.8} \\
\midrule
TED \cite{ding2024joint} & 14.1  & 87.4  & 36.6  & 72.4  & 21.3  & 61.1  & 66.1  & 56.9  & 72.5  & 69.9  & 75.4  \\
\rowcolor[rgb]{ .851,  .851,  .851} TED + Ours &       & \textbf{88.3} & \textbf{38.5} & \textbf{73.6} & \textbf{23.4} & \textbf{63.6} & \textbf{67.9} & \textbf{58.6} & \textbf{73.9} & \textbf{71.8} & \textbf{76.0} \\
\midrule
A2Net \cite{li2023lightweight} & 3.93  & 87.8  & 37.4  & 72.8  & 22.3  & 61.9  & 66.7  & 57.4  & 72.9  & 69.3  & 76.9  \\
\rowcolor[rgb]{ .851,  .851,  .851} A2Net + Ours &       & \textbf{88.1} & \textbf{38.3} & \textbf{73.3} & \textbf{23.2} & \textbf{63.4} & \textbf{67.5} & \textbf{58.3} & \textbf{73.6} & \textbf{72.4} & \textbf{76.9} \\
\midrule
SCanNet \cite{ding2024joint} & 17.81 & 87.8  & 37.8  & 72.9  & 22.8  & 62.6  & 66.8  & 57.7  & 73.1  & 70.6 & 75.9  \\
\rowcolor[rgb]{ .851,  .851,  .851} SCanNet + Ours &       & \textbf{89.0} & \textbf{39.2} & \textbf{73.7} & \textbf{24.4} & \textbf{65.1} & \textbf{67.9} & \textbf{58.3} & \textbf{74.1} & \textbf{72.6}  & \textbf{76.0} \\
\bottomrule
\end{tabular}}
 \setlength{\abovecaptionskip}{0.05cm}
  \caption{Performance of our ChangeDiff using as data augmentation on the \textit{SECOND} dataset. The training data of ChangeDiff comes from 5\%, 20\%, and 100\% of the training samples, respectively. The \textcolor[rgb]{ .749,  .561,  0}{metric} is used for semantic CD, and the \textcolor[rgb]{ .184,  .459,  .71}{metric} is used in binary CD.}
  \label{SECOND_comp}
\end{table}

\begin{table}[t]
  \centering
   \renewcommand{\arraystretch}{1.05}
  \resizebox{\linewidth}{!}{
\begin{tabular}{cccccccccccc}
\toprule
\multicolumn{12}{c}{Train on 5\% Landsat-SCD Train Set} \\
\midrule
Methods & Params. (M)  & \textcolor[rgb]{ .749,  .561,  0}{OA} & \textcolor[rgb]{ .749,  .561,  0}{Score} & \textcolor[rgb]{ .749,  .561,  0}{mIoU} & \textcolor[rgb]{ .749,  .561,  0}{Sek} & \textcolor[rgb]{ .749,  .561,  0}{$\mathrm{F_{scd}}$} & \textcolor[rgb]{ .749,  .561,  0}{Kappa} & \textcolor[rgb]{ .184,  .459,  .71}{IoU} & \textcolor[rgb]{ .184,  .459,  .71}{F1} & \textcolor[rgb]{ .184,  .459,  .71}{Rec.} & \textcolor[rgb]{ .184,  .459,  .71}{Pre.} \\
\midrule
SSCDL \cite{ding2022bi} & 13.22 & 45.4  & 24.7  & 41.4  & 20.4  & 39.3  & 38.1  & 35.7  & 42.0  & 41.6  & 42.4  \\
\rowcolor[rgb]{ .851,  .851,  .851} SSCDL + Ours &       & \textbf{46.6} & \textbf{26.0} & \textbf{44.2} & \textbf{22.5} & \textbf{41.5} & \textbf{40.3} & \textbf{38.8} & \textbf{45.0} & \textbf{43.3} & \textbf{44.1} \\
\midrule
BiSRNet \cite{ding2022bi} & 13.31 & \textbf{41.8}  & \textbf{30.8}  & 36.7  & 26.1  & 37.9  & 36.8  & \textbf{36.3}  & 40.8  & 41.3  & 40.4  \\
\rowcolor[rgb]{ .851,  .851,  .851} BiSRNet + Ours &       & 41.5 & 27.9 & \textbf{39.1} & \textbf{29.4} & \textbf{42.0} & \textbf{38.6} & 33.1 & \textbf{43.0} & \textbf{44.1} & \textbf{42.1} \\
\midrule
TED  \cite{ding2024joint} & 14.1  & 44.3  & \textbf{32.3}  & \textbf{43.6}  & 24.6  & 41.8  & \textbf{43.9}  & \textbf{40.8}  & \textbf{46.4}  & \textbf{46.5}  & 46.2  \\
\rowcolor[rgb]{ .851,  .851,  .851} TED+Ours &     & \textbf{49.1} & 28.5 & 43.4 & \textbf{26.4} & \textbf{43.7} & 39.8 & 36.9 & 46.2 & 45.8 & \textbf{46.6} \\
\midrule
A2Net \cite{li2023lightweight} & 3.93  & 38.6  & 25.6  & 35.2  & 27.2  & \textbf{39.4}  & 36.1  & 31.7  & 39.5  & 40.5  & 38.5  \\
\rowcolor[rgb]{ .851,  .851,  .851} A2Net + Ours &       & \textbf{44.3} & \textbf{33.7} & \textbf{39.3} & \textbf{28.2} & 39.3 & \textbf{39.6} & \textbf{37.9} & \textbf{43.4} & \textbf{43.1} & \textbf{43.6} \\
\midrule
SCanNet \cite{ding2024joint} & 17.81 & 47.7  & 36.8  & 45.6  & 30.5  & 45.2  & 45.2  & 43.0  & 45.2  & 45.5  & 45.0  \\
\rowcolor[rgb]{ .851,  .851,  .851} SCanNet + Ours &       & \textbf{50.1} & \textbf{38.7} & \textbf{47.7} & \textbf{33.2} & \textbf{47.5} & \textbf{47.7} & \textbf{44.2} & \textbf{46.5} & \textbf{46.9} & \textbf{46.2} \\
\midrule
\multicolumn{12}{c}{Train on 100\% Landsat-SCD Train Set} \\
\midrule
Methods & Params. (M)  & \textcolor[rgb]{ .749,  .561,  0}{OA} & \textcolor[rgb]{ .749,  .561,  0}{Score} & \textcolor[rgb]{ .749,  .561,  0}{mIoU} & \textcolor[rgb]{ .749,  .561,  0}{Sek} & \textcolor[rgb]{ .749,  .561,  0}{$\mathrm{F_{scd}}$} & \textcolor[rgb]{ .749,  .561,  0}{Kappa} & \textcolor[rgb]{ .184,  .459,  .71}{IoU} & \textcolor[rgb]{ .184,  .459,  .71}{F1} & \textcolor[rgb]{ .184,  .459,  .71}{Rec.} & \textcolor[rgb]{ .184,  .459,  .71}{Pre.} \\
\midrule
SSCDL \cite{ding2022bi} & 13.22 & 94.4  & 57.7  & 84.2  & 46.3  & 82.7  & 82.2  & 74.5  & 85.4  & 85.8  & 85.0  \\
\rowcolor[rgb]{ .851,  .851,  .851} SSCDL + Ours &       & \textbf{96.8} & \textbf{60.1} & \textbf{86.7} & \textbf{48.3} & \textbf{83.0} & \textbf{84.0} & \textbf{75.6} & \textbf{86.5} & \textbf{87.9} & \textbf{85.2} \\
\midrule
BiSRNet \cite{ding2022bi} & 13.31 & 94.5  & 58.3  & 84.3  & 54.2  & 83.4  & 82.4  & 74.7  & 85.5  & 86.1  & 84.9  \\
\rowcolor[rgb]{ .851,  .851,  .851} BiSRNet + Ours &       & \textbf{95.0} & \textbf{59.2} & \textbf{86.3} & \textbf{56.5} & \textbf{85.7} & \textbf{83.4} & \textbf{76.9} & \textbf{87.2} & \textbf{88.8} & \textbf{85.6} \\
\midrule
TED \cite{ding2024joint} & 14.1  & 95.9  & 66.5  & 88.2  & 57.2  & 87.5  & 87.2  & 80.9  & 89.5  & 89.5  & 89.4  \\
\rowcolor[rgb]{ .851,  .851,  .851} TED + Ours &       & \textbf{98.3} & \textbf{66.5} & \textbf{89.1} & \textbf{58.0} & \textbf{87.7} & \textbf{87.2} & \textbf{81.5} & \textbf{90.6} & \textbf{90.3} & \textbf{91.0} \\
\midrule
A2Net \cite{li2023lightweight} & 3.93  & 94.4  & 57.9  & 84.1  & 46.7  & 83.1  & 82.2  & 74.4  & 85.3  & 86.0  & 84.7  \\
\rowcolor[rgb]{ .851,  .851,  .851} A2Net + Ours &       & \textbf{94.5} & \textbf{58.5} & \textbf{85.4} & \textbf{46.8} & \textbf{84.7} & \textbf{83.7} & \textbf{75.5} & \textbf{86.4} & \textbf{87.6} & \textbf{85.3} \\
\midrule
SCanNet \cite{ding2024joint} & 17.81 & 96.5  & 69.9  & 89.4  & 61.5  & 89.6  & 88.6  & 82.8  & \textbf{90.6}  & 91.0  & 90.2  \\
\rowcolor[rgb]{ .851,  .851,  .851} SCanNet + Ours &       & \textbf{97.6} & \textbf{70.5} & \textbf{90.4} & \textbf{63.6} & \textbf{91.2} & \textbf{90.2} & \textbf{84.2} & 90.1 & \textbf{92.0} & \textbf{91.3} \\
\bottomrule
\end{tabular}}
 \setlength{\abovecaptionskip}{0.05cm}
  \caption{Performance of our ChangeDiff using as data augmentation on the \textit{Landsat-SCD} dataset. 
  %The training data of ChangeDiff comes from 5\%, and 100\% of the training samples, respectively.
}
  \label{Landsat-SCD_comp}
  %\vspace{-0.1cm}
\end{table}

\section{Experiments}
\subsection{Datasets and Experimental Setup}
\noindent \textbf{Setup.}
We evaluate the effectiveness of ChangeDiff on semantic change detection tasks using the following two settings:
\textbf{\textit{1) Data Augmentation}}: This setup aims to verify if synthetic data from ChangeDiff can enhance the model's discrimination capability on in-domain samples. We use three commonly used semantic change detection datasets, including SECOND \cite{yang2021asymmetric}, Landsat-SCD \cite{yuan2022transformer}, and HRSCD \cite{daudt2019multitask}. We train our ChangeDiff on these datasets, respectively. \textbf{\textit{2) Pre-training Transfer}}: This setup aims to verify if ChangeDiff can leverage publicly available semantic segmentation data to synthesize extensive data for pre-training, benefiting downstream change detection tasks. 
We use additional semantic segmentation data from LoveDA \cite{wang2021loveda} as the training source for ChangeDiff and perform pre-training with the synthetic data. We then validate its effectiveness on the SECOND and HRSCD target datasets using two transfer ways, including ``zero-shot transfer'' and ``fine-tuning transfer''. 
%``fine-tuning transfer'' refers to testing with categories that are similar between source and target datasets, while ``direct transfer'' involves adjusting the pre-trained model using 1\%, 5\%, and 100\% of the target data.

\noindent \textbf{Datasets.} Please refer to the \textcolor{red!80!black}{Appendix B}.

\noindent \textbf{Implementation Details.} Please refer to the \textcolor{red!80!black}{Appendix C}.

\subsection{Data Augmentation:}
We validate the effectiveness of ChangeDiff as data augmentation on three datasets: SECOND, Landsat-SCD, and HRSCD. 
SECOND and Landsat-SCD datasets contain incomplete (sparse) semantic annotations, while HRSCD has complete semantic annotations. 
We integrate ChangeDiff with various semantic change detection methods, including CNN-based approaches like SSCDL, BiSRNet, TED, the lightweight A2Net, and the Transformer-based SCanNet. 
Experimental results clearly demonstrate the effectiveness of our method, particularly in addressing semantic imbalance (measured by the Sek metric) and improving binary change detection performance (measured by the F1-score).

\noindent \textit{\textbf{Augmentation for SECOND Dataset.}}
As shown in Table~\ref{SECOND_comp}, for models trained on just 5\% of the SECOND dataset, Integrating our method with existing approaches consistently enhances their performance, with average improvements of 2.5\% in SeK, 2.3\% in IoU, 2.1\% in F1 score, notable gains in recall, and precision, demonstrating the effectiveness of our method across various models.
As the training set scales to 20\% and 100\% of the data, our method continues to deliver substantial improvements. 
Integrating our method with existing models yields significant improvements, with average gains of 2.4\% in SeK, 1.7\% in IoU, and 1.4\% in F1 score, as well as notable enhancements in recall and precision, showing its effectiveness across different training scenarios.

\noindent \textit{\textbf{Augmentation for Landsat-SCD Dataset.}} 
The low resolution of the Landsat-SCD dataset presents a greater challenge for effective data synthesis.
In Table \ref{Landsat-SCD_comp}, with 5\% training samples, ChangeDiff improves SeK across all models.
SeK increased from 20.4\% to 22.5\% with SSCDL and from 26.1\% to 32.4\% with BiSRNet. Similarly, F1-scores also see significant improvements, with SSCDL's F1 rising from 42.0\% to 45.0\% and A2Net's from 39.5\% to 43.4\%. With 100\% training samples, SeK continued to improve, reflecting enhanced performance on imbalanced data, such as SSCDL’s SeK increasing from 46.3\% to 48.3\%. F1 scores further improved, with TED’s F1 rising from 89.5\% to 90.6\%.

\noindent \textit{\textbf{Augmentation for HRSCD Dataset.}} 
HRSCD dataset suffers from class imbalance problems and its annotations are relatively coarse. Despite this, our ChangeDiff still shows a stable performance improvement. Specifically,
with 5\% training samples, the method enhanced SeK and F1 across all models, indicating better handling of semantic imbalance and improved detection accuracy. With 100\% training samples, ChangeDiff further boosted SeK and F1, highlighting its capability to refine both imbalance management and overall performance in change detection.

\begin{table}[tbp]
  \centering
   \renewcommand{\arraystretch}{1.05}
  \resizebox{\linewidth}{!}{
\begin{tabular}{cccccccccccc}
\toprule
\multicolumn{12}{c}{Train on 5\% HRSCD Train Set} \\
\midrule
Methods & Params. (M) & \textcolor[rgb]{ .749,  .561,  0}{OA} & \textcolor[rgb]{ .749,  .561,  0}{Score} & \textcolor[rgb]{ .749,  .561,  0}{mIoU} & \textcolor[rgb]{ .749,  .561,  0}{Sek} & \textcolor[rgb]{ .749,  .561,  0}{$\mathrm{F_{scd}}$} & \textcolor[rgb]{ .749,  .561,  0}{Kappa} & \textcolor[rgb]{ .184,  .459,  .71}{IoU} & \textcolor[rgb]{ .184,  .459,  .71}{F1} & \textcolor[rgb]{ .184,  .459,  .71}{Rec.} & \textcolor[rgb]{ .184,  .459,  .71}{Pre.} \\
\midrule
BiSRNet \cite{ding2022bi} & 13.31 & 41.1  & 18.2  & 36.3  & 12.2  & 30.9  & \textbf{33.5}  & 28.1  & \textbf{37.3}  & \textbf{37.3}  & \textbf{37.3}  \\
\rowcolor[rgb]{ .851,  .851,  .851} BiSRNet + Ours &       & \textbf{43.5} & \textbf{20.9} & \textbf{37.2} & \textbf{14.1} & \textbf{31.9} & 32.8 & \textbf{29.9} & 36.7 & 36.2 & 37.1 \\
\midrule
TED \cite{ding2024joint} & 14.1  & 41.3  & 19.0  & 35.5  & 12.1  & 28.8  & 32.0  & 28.2  & 35.1  & 34.3  & 35.9  \\
\rowcolor[rgb]{ .851,  .851,  .851} TED + Ours &       & \textbf{44.2} & \textbf{20.8} & \textbf{37.9} & \textbf{14.9} & \textbf{29.0} & \textbf{34.0} & \textbf{30.6} & \textbf{36.6} & \textbf{34.6} & \textbf{38.9} \\
\midrule
A2Net \cite{li2023lightweight} & 3.93  & 41.2  & 18.3  & 35.4  & 10.9  & 28.5  & 31.2  & 27.9  & 34.2  & 33.6  & 34.9  \\
\rowcolor[rgb]{ .851,  .851,  .851} A2Net + Ours &       & \textbf{45.3} & \textbf{20.9} & \textbf{36.0} & \textbf{12.9} & \textbf{30.9} & \textbf{33.9} & \textbf{29.7} & \textbf{35.7} & \textbf{33.8} & \textbf{37.8} \\
\midrule
SCanNet \cite{ding2024joint} & 17.81 & 43.2  & 19.9  & 36.0  & 13.0  & 29.2  & 32.9  & 28.9  & 35.7  & 35.2  & 36.2  \\
\rowcolor[rgb]{ .851,  .851,  .851} SCanNet + Ours &       & \textbf{47.4} & \textbf{20.7} & \textbf{36.4} & \textbf{14.3} & \textbf{29.4} & \textbf{34.2} & \textbf{29.2} & \textbf{36.7} & \textbf{36.6} & \textbf{36.7} \\
\midrule
\multicolumn{12}{c}{Train on 100\% HRSCD Train Set} \\
\midrule
Methods & Params. (M)  & \textcolor[rgb]{ .749,  .561,  0}{OA} & \textcolor[rgb]{ .749,  .561,  0}{Score} & \textcolor[rgb]{ .749,  .561,  0}{mIoU} & \textcolor[rgb]{ .749,  .561,  0}{Sek} & \textcolor[rgb]{ .749,  .561,  0}{$\mathrm{F_{scd}}$} & \textcolor[rgb]{ .749,  .561,  0}{Kappa} & \textcolor[rgb]{ .184,  .459,  .71}{IoU} & \textcolor[rgb]{ .184,  .459,  .71}{F1} & \textcolor[rgb]{ .184,  .459,  .71}{Rec.} & \textcolor[rgb]{ .184,  .459,  .71}{Pre.} \\
\midrule
BiSRNet \cite{ding2022bi} & 13.31 & 87.2  & 37.0  & 72.6  & 23.6  & 61.9  & 67.4  & 57.7  & 74.1  & 72.5  & 75.7  \\
\rowcolor[rgb]{ .851,  .851,  .851} BiSRNet + Ours &       & \textbf{89.8} & \textbf{38.0} & \textbf{75.4} & \textbf{23.9} & \textbf{64.0} & \textbf{68.9} & \textbf{60.4} & \textbf{76.5} & \textbf{75.7} & \textbf{77.4} \\
\midrule
TED \cite{ding2024joint} & 14.1  & 87.4  & 37.6  & 72.9  & 23.7  & 62.0  & 67.6  & 58.5  & 73.7  & 72.3  & 75.2  \\
\rowcolor[rgb]{ .851,  .851,  .851} TED + Ours &       & \textbf{88.8} & \textbf{39.0} & \textbf{74.1} & \textbf{25.9} & \textbf{64.0} & \textbf{68.4} & \textbf{59.8} & \textbf{75.2} & \textbf{74.4} & \textbf{76.1} \\
\midrule
A2Net \cite{li2023lightweight} & 3.93  & 87.9  & 38.0  & 73.2  & 23.8  & 62.8  & 67.9  & 58.6  & 74.2  & 72.7  & 75.8  \\
\rowcolor[rgb]{ .851,  .851,  .851} A2Net + Ours &       & \textbf{89.0} & \textbf{39.3} & \textbf{74.7} & \textbf{25.1} & \textbf{64.2} & \textbf{69.1} & \textbf{60.6} & \textbf{75.5} & \textbf{74.1} & \textbf{77.0} \\
\midrule
SCanNet \cite{ding2024joint} & 17.81 & 89.1  & 39.8  & 74.2  & 25.1  & 65.3  & 69.3  & 61.2  & 76.1  & 74.1  & 78.2  \\
\rowcolor[rgb]{ .851,  .851,  .851} SCanNet + Ours &       & \textbf{90.4} & \textbf{41.2} & \textbf{76.0} & \textbf{26.7} & \textbf{66.6} & \textbf{71.2} & \textbf{62.2} & \textbf{77.2} & \textbf{75.7} & \textbf{78.9} \\
\bottomrule
\end{tabular}}
 \setlength{\abovecaptionskip}{0.05cm}
  \caption{Performance of our ChangeDiff using as data augmentation on the \textit{HRSCD} dataset. 
  %The training data of ChangeDiff comes from 5\%, and 100\% of the training samples, respectively.
}
  \label{HRSCD}
  \vspace{-0.3cm}
\end{table} 

\begin{figure}[tbp]
    \begin{center}
    \centering 
\includegraphics[width=0.43\textwidth]{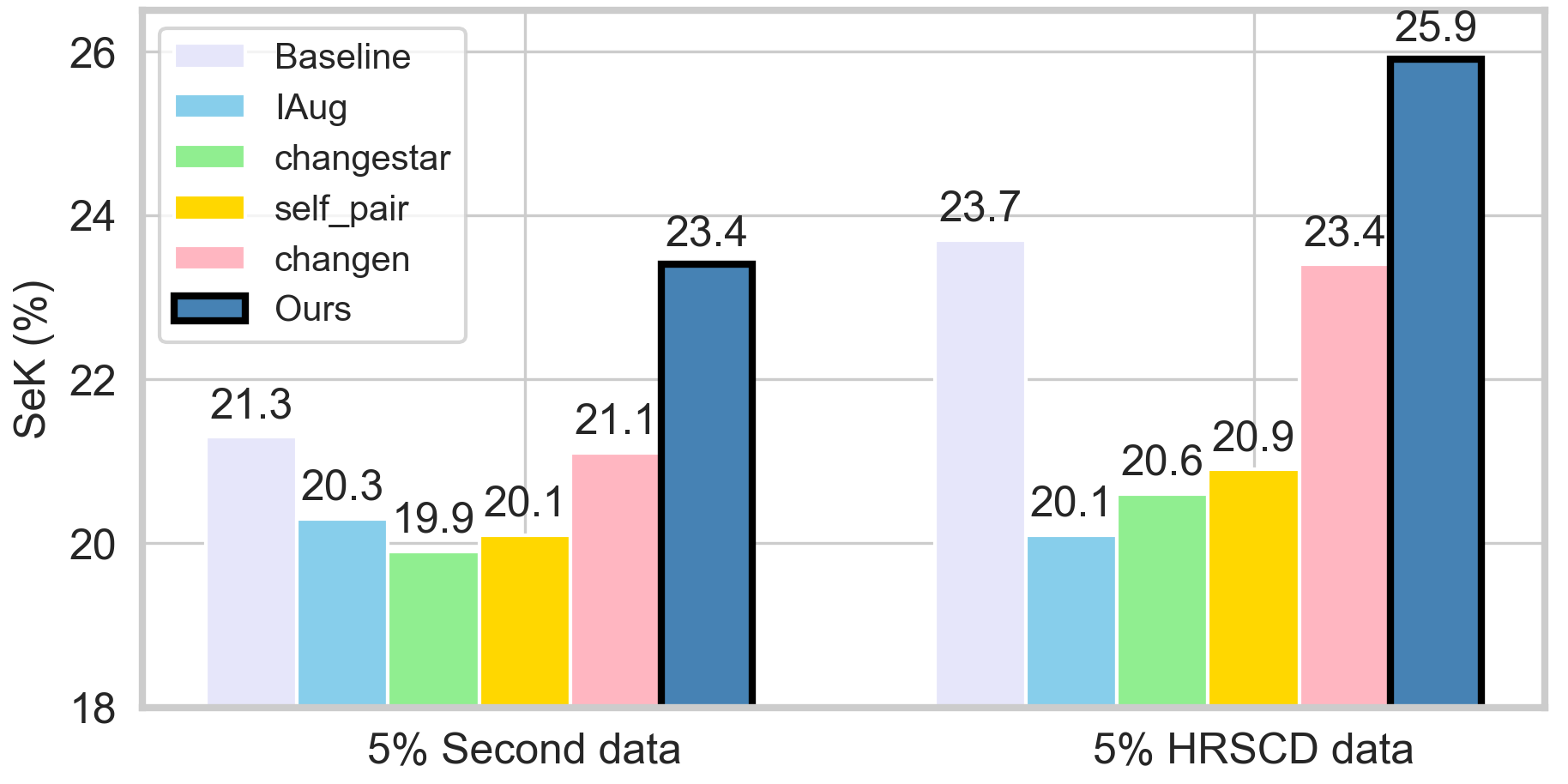}
    \end{center}
\setlength{\abovecaptionskip}{-0.15 cm}
   \caption{Comparison of data augmentation methods on two SCD datasets against competing approaches. We reproduce the results of Changen on the two SCD datasets, not using other segmentation datasets.}
    \label{compare_performance}
%\vspace{-0.45cm}
\end{figure}

\begin{figure*}[tbp]
    \begin{center}
    \centering 
\includegraphics[width=0.995\textwidth]{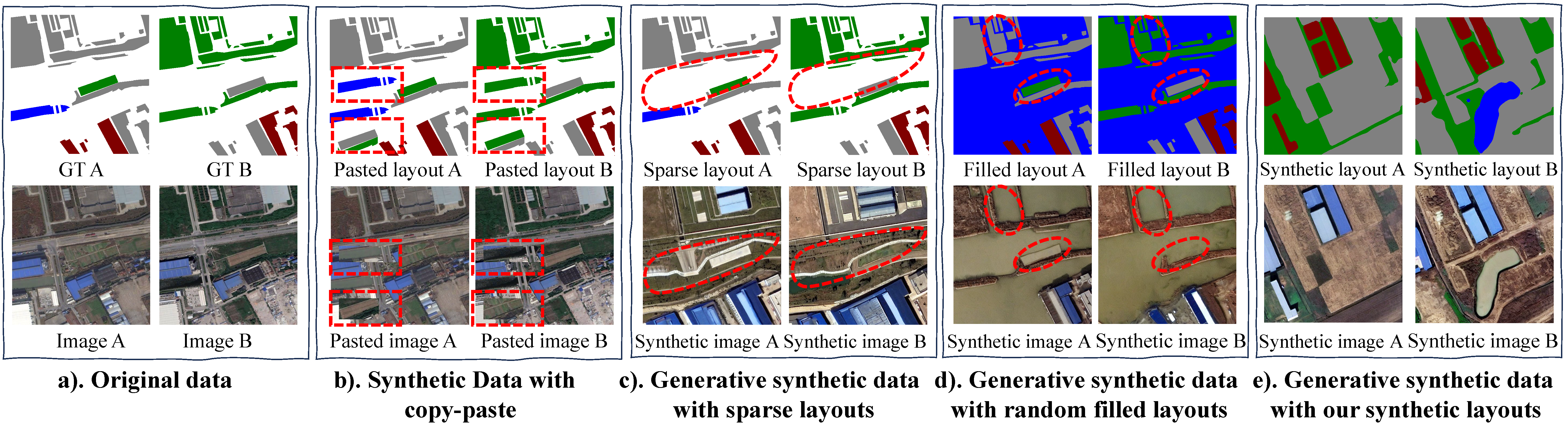}
    \end{center}
\setlength{\abovecaptionskip}{-0.15 cm}
%    \captionsetup{font={small}}
    \caption{Visual comparison of different augmentation methods for semantic change detection. 
    a) Training data in the SECOND dataset. The white areas in GT are not labeled. 
    b) Self-Pair \cite{seo2023self}: Pasting objects from other patches onto the target image to create changes. c) Changen \cite{zheng2023scalable}: Synthesizing images from a given sparse-labeled layout, and d) synthesizing images from a randomly filled layout.
    e) Ours: Synthesizing images from our synthetic layout.
    The red boxes in b), c), and d) show the low image quality and unknown semantics due to the lack of reasonable layout.}
    \label{comp_layout}
%\vspace{-0.35cm}
\end{figure*}

\noindent \textit{\textbf{Comparison with Augmentation Competitors.}} In Fig.~\ref{compare_performance}, we evaluate several data augmentation methods for change detection on the SECOND and HRSCD datasets. ChangeStar \cite{zheng2021change} and Self-Pair \cite{seo2023self}, which utilize copy-paste operations, potentially compromise image authenticity, leading to performance drops of 1.4\% SeK and 1.2\% SeK points on SECOND and HRSCD, respectively. 
IAug, based on GANs, underperforms the baseline by 1.0\% SeK on SECOND and a notable 3.6\% SeK on HRSCD. 
Changen \cite{zheng2023scalable}, a more advanced GAN-based method, shows marginal improvement on HRSCD but falls short on SECOND. In contrast, our approach surpasses the baseline by 2.1\% SeK points on SECOND and 2.2\% SeK points on HRSCD, demonstrating superior effectiveness in enhancing detection performance across diverse datasets.

\subsection{Pre-training Transfer}
In this section, we aim to validate the benefit of using ChangeDiff for data synthesis with the out-of-domain semantic segmentation dataset, LoveDA, in the context of the SCD task. We conduct experiments to validate this from two perspectives: \textit{zero-shot transfer} and \textit{fine-tuning transfer}.
More transfer experiments are in \textcolor{red!80!black}{Appendix D}.

 \begin{table}[tbp]
  \centering
   \renewcommand{\arraystretch}{1.05}
  \resizebox{\linewidth}{!}{
\begin{tabular}{c|c|ccc|ccc}
\toprule
\multicolumn{8}{c}{\textbf{\textit{Zero-shot Transfer: Train on LoveDA and Test on SECOND}}$_{test}$} \\
\hline
      &       & \multicolumn{3}{c|}{SCD} & \multicolumn{3}{c}{BCD} \\
Methods & Backbone & SeK$_{5}$ & Kappa$_{5}$ & mIoU$_{5}$ & F1 & Pre.     & Rec. \\
\hline
Copy-Paste (CP) & R-18   & 4.7   & 39.6  & 39.9  & 42.7  & 46.1  & 39.7  \\
ControlNet \cite{zhang2023adding} + CP & R-18   & 10.7  & 48.7  & 55.1  & 49.4  & 49.8  & 49.0  \\
Changen \cite{zheng2023scalable} & R-18   & 7.9   & 47.9  & 53.1  & 47.2  & 46.9  & 47.6  \\
\hline
\rowcolor[rgb]{ .851,  .851,  .851} ChangeDiff (Ours) & R-18   & \textbf{13.6} & \textbf{55.9} & \textbf{60.1} & \textbf{55.5} & \textbf{55.1} & \textbf{55.9} \\
\bottomrule
\end{tabular}}
 \setlength{\abovecaptionskip}{0.05 cm}
  \caption{Comparison of zero-shot transfer setting. We evaluate the shared five semantic categories from LoveDA to SECOND: \textit{Barren} $\rightarrow$ \textit{ground}, \textit{Forest} $\rightarrow$ \textit{Tree}, \textit{Agriculuture} $\rightarrow$ \textit{Low Vegetation}, \textit{Water} $\rightarrow$ \textit{Water},	\textit{Building} $\rightarrow$ \textit{Building}. The SCD model used here is SCanNet.}
  \label{zero_show_trans}
  \vspace{-0.3cm}
\end{table} 

\begin{figure}[!t]
    \begin{center}
    \centering 
\includegraphics[width=0.47\textwidth]{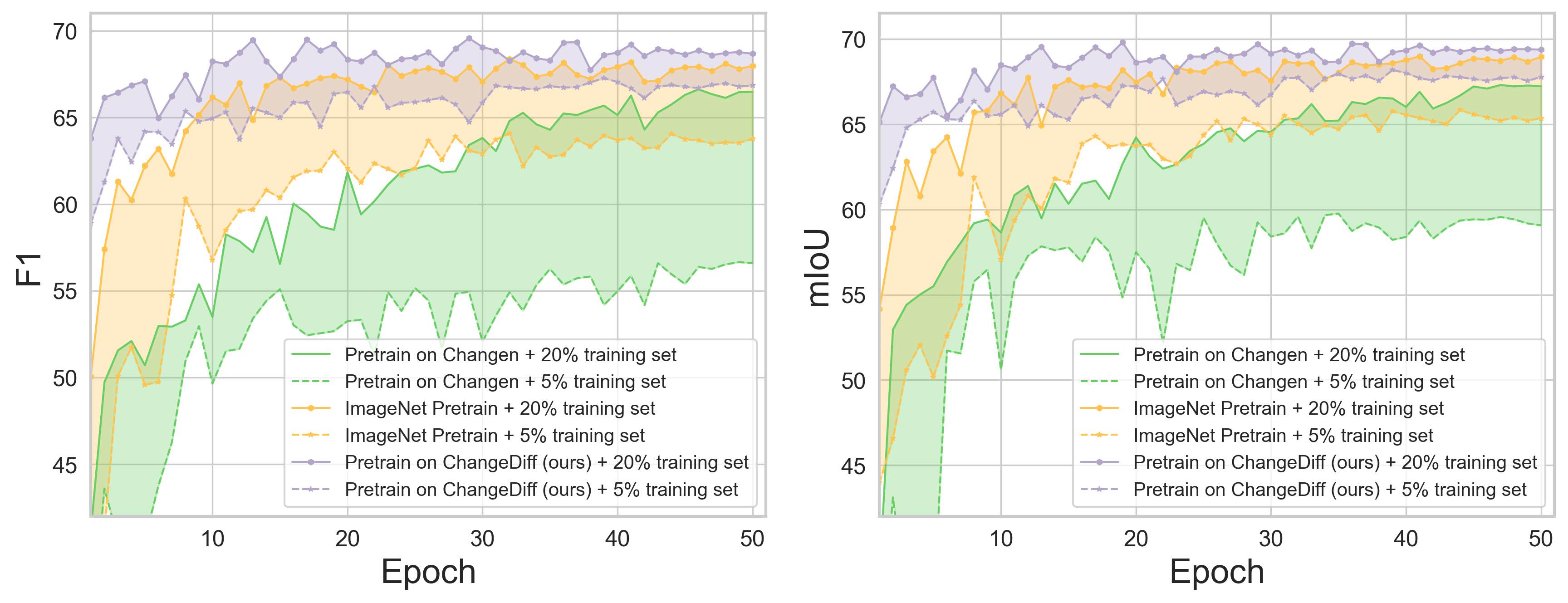}
    \end{center}
\setlength{\abovecaptionskip}{-0.15 cm}
%    \captionsetup{font={small}}
    \caption{Impact of different pre-train on model performance in the SECOND dataset. We compare: 1) pre-training with ImageNet classification data, 2) pre-training with Changen \cite{zheng2023scalable} binary synthetic change detection data \textit{(90k pairs)}, and 3) pre-training with our multi-class synthetic change detection data \textit{(10k pairs)}. They all use SCanNet with ResNet-18 as the backbone.}
    \label{compare_cuve}
\vspace{-0.2cm}
\end{figure}

\noindent \textit{\textbf{Zero-shot Transfer.}} As shown in Table~\ref{zero_show_trans}, we evaluate the performance of several data synthesis methods: Copy-Paste, ControlNet + Copy-Paste, Changen, and our ChangeDiff. 
These methods all synthesize an equal number of 10k images and are trained with the same model with same iterations. 
The results highlight that the ChangeDiff outperforms the others across all metrics.
Copy-Paste shows the weakest performance, with a mIoU$_{5}$ of 39.9\%, F1-score of 42.7\%, and SeK$_{5}$ of 4.7\%.
ControlNet + Copy-Paste improves performance, achieving an mIoU$_{5}$ of 55.1\%, F1-score of 49.4\%, and SeK$_{5}$ of 10.7\%.
Changen also performs reasonably well, with a mIoU$_{5}$ of 53.1\%, F1-score of 47.2\%, and SeK$_{5}$ of 7.9\%.
ChangeDiff (Ours) demonstrates the best results, with a mIoU$_{5}$ of 60.1\%, F1-score of 55.5\%, and SeK$_{5}$ of 13.6\%.
These results suggest that our ChangeDiff generates higher-quality synthetic data with better transferability to downstream tasks.

\noindent \textit{\textbf{Fine-tuning Transfer.}}
%We plot the performance curves of various pre-trained models on the validation set as the training progresses in Fig.~\ref{compare_cuve}. The curves show the SCD metric (right) and the BCD metric (left). Our ChangeDiff, pretrained on 10k pair of synthetic change data with 8 semantic classes from LoveDA, demonstrates faster convergence and higher average and peak accuracy across different training sample ratios. Compared to ImageNet pretraining, our ChangeDiff significantly accelerates model convergence and reduces the accuracy gap between models trained with 5\% and 20\% of the training set. Furthermore, ChangeDiff notably improves model performance. In contrast, Changen's binary classification pretraining performs poorly on multiclass tasks, despite using more synthetic data. This is attributed to Changen's 90k pretraining data focusing primarily on change in building targets, whereas the target data in SECOND includes diverse change types, leading to suboptimal guidance from Changen's pretraining.
We plot the performance curves of various pre-trained models on the validation set in Fig.~\ref{compare_cuve}, showing the SCD (right) and BCD (left) metrics. Our ChangeDiff, pretrained on 10k synthetic change data pairs with 8 semantic classes from LoveDA, achieves faster convergence and higher accuracy across different training sample ratios. Compared to ImageNet pretraining, ChangeDiff accelerates convergence and reduces the accuracy gap between models trained with 5\% and 20\% of the training set. In contrast, Changen’s binary classification pretraining performs poorly on multiclass tasks despite using more data, as its focus on building changes is less suitable for diverse target data in SECOND.

\noindent \textbf{Ablation Studies.}
%\subsection{Ablation Studies}
We ablate our core module MCDG-TP on three experiments: augmentation for SECOND and Landsat-SCD, and fine-tuning transfer on SECOND. 
We replace MCDG-TP with two strong variants: Copy-Paste and original T2I, and present the performance results in Table. \ref{alb}.
The ablation studies highlight the significant effectiveness of our MCDG-TP module. 
When compared to the baseline, MCDG-TP improves SeK by 2.3\% and F1 by 1.3\% in the Augment SECOND scenario. In the Augment Landsat-SCD scenario, shows enhancements of 2.4\% in SeK and 3.0\% in F1. For Fine-tuning Transfer, MCDG-TP achieves improvements of 2.7\% in SeK and 1.3\% in F1. These results demonstrate that MCDG-TP consistently delivers substantial performance gains across different experimental setups, outperforming the alternative methods.

\begin{table}[tbp]
  \centering
   \renewcommand{\arraystretch}{1.05}
  \resizebox{\linewidth}{!}{
\begin{tabular}{r|cccc|cc|cc}
\toprule
\multicolumn{1}{c|}{\multirow{3}[2]{*}{Methods}} & \multicolumn{4}{c|}{Augment SECOND} & \multicolumn{2}{c|}{Augment Landsat-SCD} & \multicolumn{2}{c}{Fine-tuning Transfer} \\
      & \multicolumn{2}{c}{5\%} & \multicolumn{2}{c|}{100\%} & 5\%   & 100\% & \multicolumn{2}{c}{100\%} \\
      & SeK   & F1    & SeK   & F1    & SeK   & F1    & SeK   & F1 \\
\hline
\multicolumn{1}{c|}{Baseline} & 11.3  & 64.0  & 22.8  & 73.1  & 30.5  & 45.2  & 22.8  & 73.1  \\
\hline
\multicolumn{1}{c|}{Copy-Paste + L2I} & 10.6  & 62.1  & 21.5  & 71.1  & 28.6  & 41.6  & 22.9  & 72.7  \\
\multicolumn{1}{c|}{Original T2L + L2I} & 11.1  & 63.1  & 20.8  & 71.7  & 28.1  & 42.3  & 23.2  & 72.1  \\
\hline
\multirow{2}[2]{*}{\tabincell{c}{MCDG-TP + \\T2L (Ours) + L2I}} & 13.6  & 65.3  & 24.4  & 74.1  & 33.2  & 46.5  & 24.9  & 74.3  \\
      & \textcolor[RGB]{18,220,168}{(+2.3)} &  \textcolor[RGB]{18,220,168}{(+1.3)} & \textcolor[RGB]{18,220,168}{(+1.6)} & \textcolor[RGB]{18,220,168}{(+1.1)} & \textcolor[RGB]{18,220,168}{(+2.7)} & \textcolor[RGB]{18,220,168}{(+1.3)} & \textcolor[RGB]{18,220,168}{(+2.1)} & \textcolor[RGB]{18,220,168}{(+1.1)} \\
\bottomrule
\end{tabular}}
 \setlength{\abovecaptionskip}{0.05 cm}
  \caption{Ablation Studies on our MCDG-TP.}
  \label{alb}
  %\vspace{-0.4cm}
\end{table}

\subsection{Qualitative Analysis}
\textbf{\textit{Comparison of Synthesis Quality.}} %As shown in Fig.~\ref{comp_layout},  we compare the synthesis semantic change detection data from different synthesis methods: b) Self-Pair \cite{seo2023self}: This method involves pasting objects from other patches onto the target image to create changes. The red box highlights inconsistencies between the foreground and background in the pasted areas, resulting in noticeable mismatches. c) Changen \cite{zheng2023scalable} (Sparse Layout): Images are synthesized based on a given sparse layout. The red box shows that due to the unknown semantics of the white areas in the layout, the synthesized regions appear semantically ambiguous and unclear. d) Changen (Random Filled Layout): Images are synthesized from a randomly filled layout. The red box indicates that the quality of some regions is degraded due to unreasonable semantic filling, resulting in visible artifacts. e) Ours: Images are synthesized using our constructed layout. Due to the reasonable design of change events and layout synthesis, the quality of the generated images improves significantly, with better consistency and clarity.
As shown in Fig.~\ref{comp_layout}, we compare semantic change detection data synthesized using different methods:
b) Self-Pair \cite{seo2023self}: Objects are pasted from other patches, creating mismatches in the foreground and background (highlighted in red).
c) Changen (Sparse Layout) \cite{zheng2023scalable}: Images are synthesized from a sparse layout, with semantically unclear regions (highlighted in red) due to unknown semantics in the white areas.
d) Changen (Random Filled Layout): Images synthesized with random filling show visible artifacts due to poor semantic consistency (highlighted in red).
e) Ours: Our method generates high-quality images with improved consistency and clarity, thanks to a well-designed layout and change events.

\section{Conclusion}
Change detection (CD) benefits from deep learning, but data collection and annotation remain costly. Existing generative methods for CD face issues with realism, scalability, and generalization. We introduce ChangeDiff, a new multi-temporal semantic CD data generator using diffusion models. ChangeDiff generates realistic images and simulates continuous changes without needing paired images or external datasets. It uses a text prompt for layout generation and a refinement loss to improve generalization. Future work could extend this approach to other CD tasks and enhance model scalability.

\section{Acknowledgments}
This work was supported by the National Natural Science Foundation of China under Grant Nos. 62271377, the National Key Research and Development Program of China under Grant Nos. 2021ZD0110400, 2021ZD0110404, the Key Research and Development Program of Shannxi (Program Nos. 2023YBGY244, 2023QCYLL28, 2024GX-ZDCYL-02-08, 2024GX-ZDCYL-02-17),  the Key Scientific Technological Innovation Research Project by Ministry of Education, the Joint Funds of the National Natural Science Foundation of China (U22B2054).

%\bigskip

\bibliography{aaai25}

\end{document}